\documentclass[letterpaper, 10 pt, conference, twoside]{ieeeconf}

\newif\ifreview
\reviewfalse           %

\IEEEoverridecommandlockouts
\usepackage{graphicx}
\usepackage{capt-of}
\usepackage{amsmath}
\usepackage{amssymb}
\usepackage{booktabs}
\usepackage{multirow}
\usepackage{xcolor}
\usepackage{colortbl}
\usepackage{bbding}
\usepackage[hidelinks]{hyperref}

\newcommand{\nymeria}{Nymeria}
\newcommand{\naturalmotion}{Natural Motion}
\newcommand{\boneshuman}{BONES-Seed (Human)}
\newcommand{\bonesgone}{BONES-Seed (G1)}
\newcommand{\bones}{BONES-Seed}
\newcommand{\superodom}{Real-Trajectory}

\newcommand{\tlio}{TLIO}
\newcommand{\airio}{AirIO}
\newcommand{\airimu}{AirIMU}
\newcommand{\tartanimu}{TartanIMU}
\newcommand{\imuclip}{IMU2CLIP}
\newcommand{\imagebind}{ImageBind}

\newcommand{\norm}[1]{\left\lVert#1\right\rVert}

\newcommand{\best}[1]{\textbf{#1}}

\definecolor{oursshade}{HTML}{f7f0de}

\definecolor{yescolor}{HTML}{7ba79d}
\definecolor{nocolor}{HTML}{dd8047}
\newcommand{\yes}{\textcolor{yescolor}{\CheckmarkBold}}
\newcommand{\no}{\textcolor{nocolor}{\XSolidBrush}}

\title{\LARGE \bf
A Unified Inertial Encoder for Human and Humanoid Motion
}

\ifreview
  \author{Anonymous Authors}
\else
  \author{
    \parbox{\dimexpr\textwidth-2\tabcolsep}{\centering\normalfont
      \vspace{-\baselineskip}%
      \raisebox{4pt}{\small\textcolor[HTML]{AA9B65}{\href{https://unified-inertial-encoder.github.io/}{\texttt{unified-inertial-encoder.github.io}}}}\\[5pt]
      {\bfseries
      Yutian Chen$^{1}$, Shibo Zhao$^{1}$, Rocky Duan$^{1}$, Karen Liu$^{1,3}$, Carmelo Sferrazza$^{1}$, \\
      Guanya Shi$^{1,4}$, Yuheng Qiu$^{1}$ and Koushil Sreenath$^{1,2}$}\\
      $^{1}$Amazon FAR, $^{2}$UC Berkeley, $^{3}$Stanford University, $^{4}$Carnegie Mellon University
      \vspace{-7pt}%
      }%
    }
\fi

\begin{document}

\IEEEaftertitletext{%
    \vspace{-0.5\baselineskip}%
    \begin{center}%
        \includegraphics[width=\textwidth]{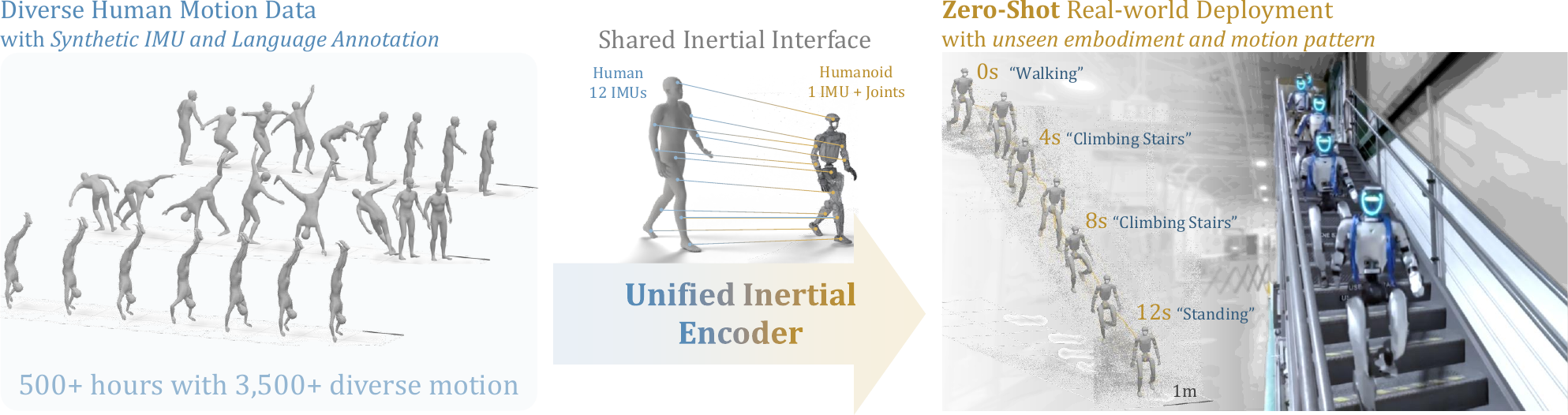}%
        \vspace{-10pt}%
        \captionof{figure}{We present a unified inertial encoder for human and humanoid motion. The encoder is trained on diverse human and IK-retargeted G1 motions (left), and uses a shared inertial interface to connect corresponding body sites across embodiments (center). Our encoder transfers to a physical G1 humanoid (right) for the inertial odometry and motion-description retrieval tasks without requiring physics simulation, real-robot trajectories, or fine-tuning. 
        }
        \label{fig:teaser}%
    \end{center}%
    \vspace{-0.25\baselineskip}}
\maketitle
\thispagestyle{empty}
\pagestyle{empty}

\begin{abstract}
Learning humanoid state estimation and language-grounded motion understanding from robot recordings ties training coverage to the robot's motion repertoire. Human motion datasets offer broader coverage, yet transferring this knowledge requires overcoming differences in body geometry, sensing, and physical disturbances. To enable this transfer, we propose a unified inertial encoder whose shared inertial interface uses kinematic IMU synthesis to represent human and humanoid motion at corresponding body sites. Within this common input space, joint training on human and kinematically retargeted humanoid motion combines velocity estimation and motion--text alignment with continuous-time encoding and sensor and motion randomization. The resulting model transfers zero-shot to a physical Unitree G1 without physics simulation, real-robot training trajectories, or fine-tuning, achieving the lowest velocity error among evaluated methods across five datasets, including 0.099~m/s on real G1. Its motion-description retrieval also extends to retargeted G1 motion categories observed only in human training, demonstrating how human motion knowledge can support the generalization of  humanoid inertial odometry and language-grounded motion understanding.
\end{abstract}

\section{INTRODUCTION}
\label{sec:introduction}

Humanoids rely on state estimation for locomotion and balance~\cite{kuindersma2016atlas,rotella2014state}, and can use language-grounded motion understanding to interpret their behavior for monitoring and human--robot interaction~\cite{hayes2017transparency}.
Learning these capabilities from robot recordings, however, ties training coverage to behaviors that existing controllers can execute~\cite{fu2024humanplus}.
For state estimation, this creates a circular dependence: reliable control requires accurate estimates, yet estimators trained on robot recordings can inherit the controller's limited motion repertoire.
Human datasets can broaden this coverage with diverse whole-body movements~\cite{ma2024nymeria,geissinger2020motion} and natural-language descriptions~\cite{ma2024nymeria}, without requiring the humanoid to perform each behavior.

Realizing this opportunity remains challenging, even though inertial measurement units (IMUs) provide a common sensing modality.
Existing learned inertial odometry typically specializes to a platform or dataset~\cite{chen2018ionet,herath2020ronin,liu2020tlio,qiu2025airio}, while reusable IMU representations primarily study human tasks~\cite{moon2023imu2clip,das2025primus}.
Extending these representations to humanoids requires reconciling differences in body geometry, sensor placement, and coordinate conventions~\cite{geissinger2020motion,he2024h2o}, along with varying sampling rates and measurement timing.
Although retargeting can supply humanoid kinematics, the synthesized signals do not capture all contact effects, structural vibration, impacts, and sensor imperfections present in physical measurements~\cite{fallon2018localization}.
A shared representation must therefore accommodate differences across embodiments while remaining robust to the gap between synthesized signals and physical IMU measurements.

Our central insight is to establish this correspondence through a shared inertial interface at anatomically corresponding body sites.
For human motion, these sites receive signals from wearable IMUs or virtual sensors derived from kinematics; for humanoids, we synthesize the corresponding signals by combining pelvis IMU and joint encoder measurements with URDF kinematics.
This kinematic IMU synthesis incorporates embodiment geometry into the input representation, which we then align to common coordinate conventions.
This input space lets a unified inertial encoder learn from human and humanoid motion.

We jointly train this encoder on human and kinematically retargeted humanoid motion for velocity estimation and motion--text alignment, so that one model supports both inertial odometry and language-grounded motion understanding.
Continuous-time encoding accommodates sampling differences across these sources, including different rates, missing samples, and irregular timing.
Sensor and motion randomization further prepares the encoder for motion variation and disturbances absent from kinematic training signals.
Together, these choices enable zero-shot transfer to a physical Unitree G1 without physics simulation, real-robot training trajectories, or hardware-specific fine-tuning.

The benefit of learning from human motion is evident in real-G1 trajectory error with ground-truth attitude, which falls from 5.901~m with humanoid-only training to 3.153~m with joint training.
The same checkpoint achieves the lowest absolute velocity error among evaluated methods across five datasets, including 0.099~m/s on real G1.
The transfer extends to language-grounded motion understanding, with motion-description retrieval reaching 53.8\% recall at 1 versus 17.3\% for IMU2CLIP on retargeted G1 categories observed only in human training.
Beyond this retargeted evaluation, demonstrations on a physical G1 illustrate motion-description retrieval on hardware.

In summary, our contributions are:
\begin{itemize}
\hyphenpenalty=1000
\item We show human motion improves humanoid inertial odometry and enables zero-shot transfer to a physical G1 without real-robot training or fine-tuning.
\item We present a shared inertial interface incorporating embodiment geometry via kinematic IMU synthesis, with continuous-time encoding across sampling rates.
\item We learn a language-aligned IMU feature space supporting motion-description retrieval without task-specific classifiers, including humanoid motion categories observed only in human training.
\end{itemize}

\section{RELATED WORK}
\label{sec:related_work}

\subsection{Learned Inertial Odometry}
Learned inertial odometry constrains drift from IMU integration by drawing on motion priors learned from data.
IONet~\cite{chen2018ionet} and RoNIN~\cite{herath2020ronin} use these priors to predict windowed displacement and pedestrian velocity, respectively.
Learned outputs also support model-based estimation: TLIO~\cite{liu2020tlio} fuses displacement and uncertainty in a filter, while AirIMU~\cite{qiu2025airimu} combines IMU corrections and uncertainty with integration.
For highly dynamic UAV motion, AirIO~\cite{qiu2025airio} uses body-frame measurements to improve velocity observability.
Across these formulations, the learned priors and corrections reflect the sensor characteristics and behaviors encountered during training, so estimation accuracy can degrade when deployment motions fall outside that distribution~\cite{qiu2025airio,zhao2025tartanimu}.
Their effectiveness in new settings thus depends in part on the breadth of training coverage.

Efforts to improve this generalization address both geometric variation and motion coverage: RIO~\cite{cao2022rio} and EqNIO~\cite{jayanth2025eqnio} use equivariance for geometric robustness, while TartanIMU~\cite{zhao2025tartanimu} combines cross-platform training with fine-tuning and online adaptation.
For humanoids, broadening coverage also raises the question of where diverse training motion can come from.
Human datasets offer behaviors beyond those available from an existing robot controller~\cite{ma2024nymeria,geissinger2020motion}, motivating our use of human and kinematically retargeted humanoid motion to train one encoder.
We examine whether this training improves humanoid inertial odometry and supports zero-shot transfer to physical G1 hardware without real-robot training or fine-tuning.

\subsection{General-Purpose and Language-Aligned IMU Models}
General-purpose IMU models seek reusable inertial features for activity recognition and motion retrieval, with LIMU-BERT~\cite{xu2021limubert} learning contextual features from unlabeled sequences and PRIMUS~\cite{das2025primus} combining self-supervised and multimodal objectives.
Although these encoders support transfer, their downstream evaluations rely on classifiers trained with task-specific labels.
Connecting the features to natural-language descriptions provides a complementary route to motion understanding, allowing descriptions to serve as retrieval queries without training a classifier for each downstream label set.

This language alignment underlies IMU2CLIP~\cite{moon2023imu2clip}, which projects human IMU signals into a joint video--text embedding space, and ImageBind~\cite{girdhar2023imagebind}, which includes IMU among six modalities in a shared representation.
SensorLM~\cite{zhang2025sensorlm} develops sensor--language representations from large-scale wearable data, while AURA-MFM~\cite{matsuishi2025aura} aligns IMU with third-person video, motion capture, and text for retrieval and activity recognition.
Their recognition and retrieval evaluations, however, leave open whether these representations transfer motion semantics to humanoids or support velocity estimation.
We address these questions through joint velocity estimation and motion--text alignment, evaluating both inertial odometry and motion-description retrieval, including humanoid categories excluded from humanoid training but represented in human training.

\subsection{Human-to-Humanoid Motion Transfer}
Using diverse human motion for humanoid learning requires reconciling differences in morphology and physical capability.
Retargeting addresses this mismatch through approaches ranging from shared latent spaces for human and humanoid kinematic structures~\cite{choi2020retargeting} to the morphology-invariant and physics-aware constraints of Human2Humanoid~\cite{huang2026human2humanoid}.
Systems such as H2O~\cite{he2024h2o}, OmniH2O~\cite{he2024omnih2o}, and HumanPlus~\cite{fu2024humanplus} use retargeted motion to train control policies in physics simulation for deployment on real humanoids.
While these systems transfer reference motions or control policies to produce behavior, our aim is to transfer human motion knowledge at the sensing level.
Our inertial encoder, trained on human and kinematically retargeted motion, uses a shared inertial interface for state estimation and language-grounded motion understanding on a physical G1, without simulation or real-robot trajectories.

\section{METHOD}
\label{sec:method}

\begin{figure*}[t]
    \centering
    \includegraphics[width=\textwidth]{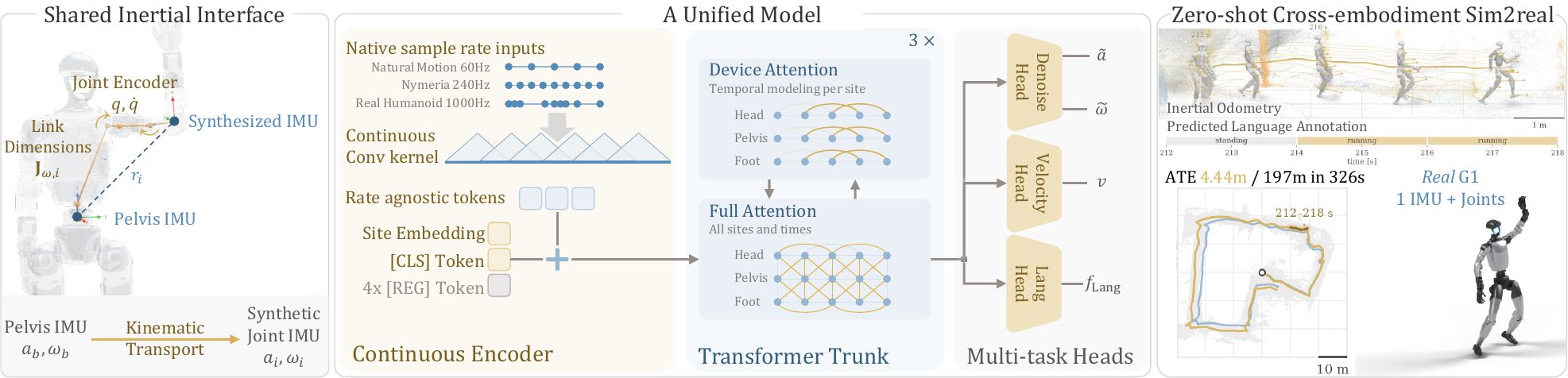}
    \vspace{-14pt}
    \caption{
    Overview of the proposed method. Left: kinematic transport combines pelvis IMU measurements, joint encoders, and link kinematics to synthesize inertial signals at anatomically corresponding sites. Center: continuous convolution converts native-rate IMU streams into fixed-duration tokens. A shared transformer alternates within-site and full attention, with heads predicting body-frame velocity, gravity-free acceleration and clean angular velocity, and a motion embedding aligned with frozen text features. Right: example outputs on a physical G1 illustrate inertial odometry and motion-description retrieval.
    }
    \label{fig:overview}
    \vspace{-16pt}
\end{figure*}

\subsection{Problem Formulation}
Let $\mathcal{S}$ denote a set of anatomically corresponding inertial sites shared by humans
and humanoids. For a motion window, site $i\in\mathcal{S}$ provides a timestamped sequence
\begin{equation}
    \mathcal{X}_i = \left\{\left(t_{i,j},\mathbf{a}_{i,j},
    \boldsymbol{\omega}_{i,j}\right)\right\}_{j=1}^{T_i},
\end{equation}
where $\mathbf{a}_{i,j},\boldsymbol{\omega}_{i,j}\in\mathbb{R}^{3}$ are measured specific
force (including the gravity contribution) and angular velocity in the local sensor frame.
The sequence length $T_i$ varies with the sensor rate and efficient inference is supported via the variable length attention~\cite{dao2023flashattention2}.
Given $\mathcal{X}=\{\mathcal{X}_i\}_{i\in\mathcal{S}}$, our objective is to learn a shared inertial representation that captures both the metric and semantic properties of motion.
For velocity estimation, the model predicts each site's body-frame linear velocity $\hat{\mathbf{v}}_{i,j}$ at its input timestamps, with pelvis velocity serving as root velocity. For semantic motion understanding, it maps the same motion window to a feature $\mathbf{f}_{\mathrm{Lang}}$ aligned with a frozen text embedding space, enabling retrieval of natural-language motion descriptions.
To support both tasks, auxiliary denoising supervision encourages the shared representation to preserve the underlying motion despite sensor disturbances by reconstructing gravity-free kinematic acceleration $\tilde{\mathbf{a}}_{i,j}$ and clean angular velocity $\tilde{\boldsymbol{\omega}}_{i,j}$.
We learn this representation jointly from human and kinematically retargeted humanoid motion, with the goal of transferring both estimation and understanding to a physical humanoid.

\subsection{Unified Multi-IMU Representation}
\label{sec:method_representation}
We first map human and humanoid motion signals into a common inertial input space.
We instantiate $\mathcal{S}$ with 12 anatomical sites: the pelvis, head, and bilateral upper arms, forearms, upper legs, lower legs, and feet, discarding streams outside this set. Human signals come from wearable IMUs or virtual sensors derived from kinematics; humanoid signals are synthesized from the pelvis IMU and joint states. We align these streams to a common coordinate convention and encode them at their native timestamps, as detailed below.

\subsubsection{Humanoid IMU synthesis}
On G1, we transport the pelvis IMU measurements to the selected links using joint positions
and velocities. Let $\mathbf{R}_{bi}$ map link-$i$ vectors into pelvis coordinates, and let
$\mathbf{r}_i$, $\dot{\mathbf{r}}_i$, and $\ddot{\mathbf{r}}_i$ denote the link origin's
position and its derivatives relative to the pelvis. Placing the base IMU at the pelvis
origin gives
\begin{equation}
\begin{aligned}
\boldsymbol{\omega}_i &= \mathbf{R}_{bi}^{\mathsf{T}}
    (\boldsymbol{\omega}_b + \mathbf{J}_{\omega,i}\dot{\mathbf{q}}),\\
\mathbf{a}_i &= \mathbf{R}_{bi}^{\mathsf{T}}\bigl(\mathbf{a}_b
    + \dot{\boldsymbol{\omega}}_b\times\mathbf{r}_i
    + \boldsymbol{\omega}_b\times(\boldsymbol{\omega}_b\times\mathbf{r}_i)\\
    &\hspace{23mm}+ 2\boldsymbol{\omega}_b\times\dot{\mathbf{r}}_i
    + \ddot{\mathbf{r}}_i\bigr).
\end{aligned}
\label{eq:imu_synthesis}
\end{equation}
Here $\boldsymbol{\omega}_b$ and $\mathbf{a}_b$ are the base IMU's angular velocity and acceleration in its sensor frame, $\mathbf{q}$ denotes joint angles, and $\mathbf{J}_{\omega,i}$ is the link's angular Jacobian in pelvis coordinates. Forward kinematics and Jacobians provide the relative link
motion; numerical differentiation of the base angular velocity and joint velocities supplies
angular and joint accelerations. The transport includes tangential, centripetal, Coriolis,
and relative-link acceleration terms, and requires no world pose or root velocity.

For retargeted G1 motion, the base signals are derived from root kinematics; on hardware,
they are read from the pelvis IMU. The synthesized accelerations inherit the gravity content
of these base signals: gravity-free kinematic inputs yield gravity-free link accelerations,
whereas measured specific force retains the gravity contribution at each link.

\subsubsection{Coordinate alignment}
Different datasets use different sensor-axis conventions, so we rotate all signals into the human Xsens convention~\cite{xsens_mvn_manual}. For each body site, we first estimate the sensor's average orientation relative to an anatomical frame constructed from joint positions. Comparing this orientation with the Xsens reference gives a fixed rotation $\mathbf{M}_i$ from the reference axes to the dataset's native axes. We then apply this to sensor orientations and signals:
\begin{equation}
    \mathbf{R}_{wi}^{\mathrm{can}}=\mathbf{R}_{wi}\mathbf{M}_i,
    \quad
    \mathbf{x}_i^{\mathrm{can}}=\mathbf{M}_i^{\mathsf{T}}\mathbf{x}_i,
    \quad \mathbf{x}_i\in\{\mathbf{a}_i,\boldsymbol{\omega}_i,\mathbf{v}_i\}.
\label{eq:mount_alignment}
\end{equation}
Here $\mathbf{R}_{wi}$ is the native sensor-to-world rotation, and $\mathrm{can}$ marks quantities after alignment. The same correction is applied to acceleration, angular velocity, and velocity targets across recordings; world positions remain unchanged. Retargeted and real G1 motion share the same correction.

\begin{table}[t]
\caption{Training augmentation for motion diversity, sampling variation, and hardware disturbances.}
\vspace{-6pt}
\label{tab:randomization}
\centering
\small
\setlength{\tabcolsep}{3pt}
\renewcommand{\arraystretch}{1.08}
\begin{tabular}{@{}p{0.35\columnwidth}p{\dimexpr0.65\columnwidth-6pt\relax}@{}}
\toprule
\textbf{Parameter} & \textbf{Setting} \\
\midrule
\multicolumn{2}{@{}l}{\textit{Motion and sampling}} \\
Gravity magnitude & $\mathcal{U}(9.76,9.83)$~m/s$^2$ \\
Body scale & $\mathcal{U}(0.75,1.0)$ / $\mathcal{U}(1.0,1.33)$ \\
Motion speed & $\mathcal{U}(0.7,1.3)$ / $\mathcal{U}(0.9,1.6)$ \\
Human sampling rate & \raggedright Target $\mathcal{U}(120,240)$~Hz \tabularnewline
Timestamp jitter & \raggedright Spacing CV target $\mathcal{U}(0,0.15)$ \tabularnewline
\midrule
\multicolumn{2}{@{}l}{\textit{Sensor noise}} \\
Acc. noise density & $10^{-2}$~(m/s$^2$)/$\sqrt{\mathrm{Hz}}$ \\
Gyro. noise density & $3\times10^{-4}$~(rad/s)/$\sqrt{\mathrm{Hz}}$ \\
\midrule
\multicolumn{2}{@{}l}{\textit{Mechanical disturbances}} \\
Vibration & Base RMS 0.3--3.0~m/s$^2$ \\
Impacts & Base peak 3--30~m/s$^2$ \\
\bottomrule
\end{tabular}
\par\smallskip
\begin{minipage}{\columnwidth}
\scriptsize
${}^\star\;\mathcal{U}$ denotes uniform distribution; paired ranges are human / G1.
CV is the coefficient of variation.
RMS and peak are values before modulation.
\end{minipage}
\vspace{-10pt}
\end{table}

\subsubsection{Continuous-time encoding and decoding}
We encode each site's inertial stream into fixed-duration tokens through a learned hierarchical continuous convolution~\cite{romero2022ckconv}.
The convolution aggregates acceleration and gyroscope measurements using weights determined by their sampling times, accommodating different sampling rates and missing samples. This avoids resampling to a fixed time grid, which can introduce interpolation artifacts and distort the signals.

At each pooling stage, let $\mathcal{J}_w$ index features $\mathbf{h}_j$ at timestamps $t_j$ within an interval starting at $\tau_w$ with duration $\Delta$.
Using normalized time $u_{w,j}=(t_j-\tau_w)/\Delta$, we compute
\begin{equation}
\begin{aligned}
\bar{\mathbf{h}}_{w,k} &=
\frac{\sum_{j\in\mathcal{J}w}\phi_k(u_{w,j})\mathbf{h}_j}
{\max \large(\epsilon,\sum_{j\in\mathcal{J}_w}\phi_k(u_{w,j})\large)}, \\
\mathbf{y}_w &= \mathbf{b}+\sum_{k=1}^{K}\mathbf{W}_k\bar{\mathbf{h}}_{w,k}.
\end{aligned}
\label{eq}
\end{equation}
Here $\phi_k(u)=1-|u-c_k|/\delta$ are linear B-spline weights with uniformly spaced centers $c_k$ and spacing $\delta$, $\mathbf{W}_k,\,\mathbf{b}$ are learned parameters, and $\epsilon>0$ is for numeric stability.

To capture both local motion and whole-body coordination, the transformer alternates attention within sites and across sites. Learned site embeddings are added once to the encoder's output to distinguish anatomical locations. Time-based rotary encoding~\cite{su2024roformer} is applied in attention to capture elapsed time between tokens. The resulting features support velocity estimation and denoising, with a {\small\texttt{[CLS]}} token summarizing the motion for language alignment.

\begin{table}[t]
\caption{Datasets used in this work. Real G1 recordings are reserved for evaluation. Emb. denotes embodiment; Freq. is in Hz; Lang. and Train mark language supervision and training use. 
}
\vspace{-6pt}
\label{tab:datasets}
\centering
\renewcommand{\yes}{\textcolor[HTML]{88A9BB}{\CheckmarkBold}}
\resizebox{\columnwidth}{!}{%
\begin{tabular}{llcccccc}
\toprule
Dataset & Emb. & Motion & Real IMU & Freq. & Hours & Lang. & Train \\
\midrule
\nymeria       & Human & Real      & \yes & 240 & 300+    & \yes & \yes \\
\naturalmotion & Human & Real      & \yes & 240 & 40.6    & \no  & \yes \\
\bones         & Human & Real      & \yes & 120 & 144     & \yes & \yes \\
\bones         & G1    & Retarget  & \yes & 120 & 144     & \yes & \yes \\
\superodom     & G1    & Real      & \no  & 200 & 0.7     & \no  & \no  \\
\bottomrule
\end{tabular}}
\vspace{-16pt}
\end{table}

\subsection{Multi-Task Joint Training}
\label{sec:method_training}
We train a single model jointly on human motion and kinematically retargeted G1 motion. The
shared trunk is optimized through four complementary objectives:
\begin{equation}
    \mathcal{L} = \lambda_{\mathrm{vel}}\mathcal{L}_{\mathrm{vel}}
    + \lambda_{\mathrm{traj}}\mathcal{L}_{\mathrm{traj}}
    + \lambda_{\mathrm{lang}}\mathcal{L}_{\mathrm{lang}}
    + \lambda_{\mathrm{den}}\mathcal{L}_{\mathrm{den}}.
\end{equation}
\subsubsection{Velocity and trajectory consistency}
The velocity loss supervises each site's predicted velocity $\hat{\mathbf{v}}$ against ground truth $\mathbf{v}$:
$\mathcal{L}_{\mathrm{vel}}=\mathbb{E}\|\hat{\mathbf{v}}-\mathbf{v}\|_2$.
To constrain accumulated drift, we also compare predicted and ground-truth displacements over short intervals. We integrate denoised angular velocity $\tilde{\boldsymbol{\omega}}$~\cite{forster2017preintegration} using fourth-order Runge--Kutta, initialized with the ground-truth attitude, to obtain the estimated sensor-to-world rotation $\hat{\mathbf{R}}$. We stop gradients through this rotation and integrate $\hat{\mathbf{R}}\hat{\mathbf{v}}$ using the trapezoidal rule to obtain world-frame position $\hat{\mathbf{p}}$. Hats denote estimates; $\mathbf{R}$ and $\mathbf{p}$ denote ground-truth rotation and position.
For an interval from sample $s$ to $e$, let $\Delta\mathbf{p}_{se}^{\mathrm{pred}}=\hat{\mathbf{R}}_s^{\mathsf{T}}(\hat{\mathbf{p}}_e-\hat{\mathbf{p}}_s)$ and $\Delta\mathbf{p}_{se}^{\mathrm{gt}}=\mathbf{R}_s^{\mathsf{T}}(\mathbf{p}_e-\mathbf{p}_s)$ denote predicted and ground-truth displacements in their respective starting sensor frames. The trajectory loss is
\begin{equation}
\mathcal{L}_{\mathrm{traj}} = \frac{1}{|\mathcal{I}|}\sum_{(s,e)\in\mathcal{I}}
\left\|\frac{\Delta\mathbf{p}_{se}^{\mathrm{pred}}-\Delta\mathbf{p}_{se}^{\mathrm{gt}}}{t_e-t_s}\right\|_2^2.
\label{eq:trajectory_loss}
\end{equation}
Here $t_s,t_e$ are the endpoint timestamps, and $\mathcal{I}$ collects intervals of available durations from $\{0.25,1,2\}$~s without overlap. Here $|\mathcal{I}|$ is the number of intervals. Dividing by elapsed time makes errors comparable across durations.

\subsubsection{Language alignment and denoising}
For both Nymeria and BONES-Seed, a window receives language supervision only if it is fully covered by a single language annotation.
A frozen CLIP ViT-L/14@336px text encoder~\cite{radford2021clip} maps the motion description to a 768-dimensional target $\mathbf{z}$, and we align the window feature $\mathbf{f}_{\mathrm{Lang}}$ using cosine distance:
\begin{equation}
    \mathcal{L}_{\mathrm{lang}} = 1-\cos(\mathbf{f}_{\mathrm{Lang}},\mathbf{z}).
\end{equation}
Windows without a valid language target still contribute to the motion and denoising objectives.

The auxiliary denoising loss supervises reconstruction of gravity-free acceleration and angular velocity derived from the underlying human or retargeted humanoid motion kinematics.
These kinematic signals provide clean targets, while gravity and simulated sensor disturbances are added to construct the model inputs.
The loss sums the mean Euclidean errors between the predicted signals and targets.

\subsubsection{Adaptive loss weights}
We fix $\lambda_{\mathrm{den}}=\lambda_{\mathrm{lang}}=1$ and adapt the velocity and trajectory weights to balance their gradient norms against denoising at a shared feature layer.
Each weight is an exponential moving average (EMA) of the denoising-to-task gradient norm ratio, clipped to $[0.02,50]$.

\begin{table*}[t]
\caption{Inertial odometry across human and humanoid datasets. AVE measures velocity error (m/s); ATE measures trajectory error (m), using ground-truth (GT) or gyro-integrated (Pred) attitude. Lower is better; bold marks column minima.}
\vspace*{-8pt}
\label{tab:odometry}
\centering
\setlength{\tabcolsep}{3pt}
\resizebox{\textwidth}{!}{%
\begin{tabular}{l ccc ccc ccc ccc ccc}
\toprule
 & \multicolumn{9}{c}{Human} & \multicolumn{6}{c}{Humanoid (G1)} \\
\cmidrule(lr){2-10}\cmidrule(lr){11-16}
 & \multicolumn{3}{c}{\nymeria} & \multicolumn{3}{c}{\naturalmotion} & \multicolumn{3}{c}{\boneshuman} & \multicolumn{3}{c}{\bonesgone} & \multicolumn{3}{c}{\superodom} \\
\cmidrule(lr){2-4}\cmidrule(lr){5-7}\cmidrule(lr){8-10}\cmidrule(lr){11-13}\cmidrule(lr){14-16}
Method & AVE & ATE\textsubscript{GT} & ATE\textsubscript{Pred} & AVE & ATE\textsubscript{GT} & ATE\textsubscript{Pred} & AVE & ATE\textsubscript{GT} & ATE\textsubscript{Pred} & AVE & ATE\textsubscript{GT} & ATE\textsubscript{Pred} & AVE & ATE\textsubscript{GT} & ATE\textsubscript{Pred} \\
\midrule
\multicolumn{16}{l}{\textit{Pelvis IMU only}} \\
\hspace{3mm}\tlio\textsuperscript{$\dagger$}~\cite{liu2020tlio} & 0.122 & 8.149 & \best{17.76} & 0.189 & 40.72 & 81.31 & 0.245 & 0.717 & 0.719 & 0.173 & 0.761 & 0.762 & 0.371 & 17.86 & 17.86 \\
\hspace{3mm}\airimu\textsuperscript{$\dagger$}~\cite{qiu2025airimu} & 12.62 & -- & -- & 109.3 & -- & -- & 0.212 & 9.472 & 505.5 & 0.281 & 11.83 & 306.6 & 6.231 & 731.1 & -- \\
\hspace{3mm}\airio\textsuperscript{$\dagger$}~\cite{qiu2025airio} & 0.069 & 7.481 & 113.9 & 0.280 & 158.1 & 157.9 & 0.140 & 0.530 & 1.918 & 0.118 & 0.720 & 1.690 & 0.350 & 16.82 & 20.48 \\
\hspace{3mm}\tartanimu~\cite{zhao2025tartanimu} & 0.301 & 37.74 & 35.53 & 0.332 & 155.8 & 156.3 & 0.328\textsuperscript{*} & 2.970\textsuperscript{*} & 2.964\textsuperscript{*} & 0.219\textsuperscript{*} & 2.176\textsuperscript{*} & 2.176\textsuperscript{*} & 0.391 & 18.94 & 18.96 \\
\midrule
\multicolumn{16}{l}{\textit{Pelvis IMU + joint positions and velocities}} \\
\hspace{3mm}\airio\ w/ joints\textsuperscript{$\dagger$} & 0.060 & 6.912 & 39.64 & 0.080 & 14.11 & 209.2 & 0.205 & 0.907 & 1.652 & 0.123 & 0.620 & 1.565 & 0.190 & 5.976 & 22.25 \\
\specialrule{\lightrulewidth}{\aboverulesep}{0pt}
\rowcolor{oursshade}
\rule[-0.9ex]{0pt}{3.1ex}Ours & \best{0.026} & \best{2.468} & 18.20 & \best{0.049} & \best{7.456} & \best{50.54} & \best{0.031} & \best{0.113} & \best{0.148} & \best{0.024} & \best{0.101} & \best{0.127} & \best{0.099} & \best{3.153} & \best{11.50} \\
\specialrule{\heavyrulewidth}{0pt}{\belowrulesep}
\end{tabular}%
}
\par\smallskip\vspace{-1mm}
\begin{minipage}{\linewidth}
\scriptsize
$\dagger\;$Trained separately on each dataset. \quad
* Only sequences longer than 10 seconds are evaluated.
\end{minipage}
\vspace{-10pt}
\end{table*}

\begin{figure*}[t]
    \centering
    \includegraphics[width=\textwidth]{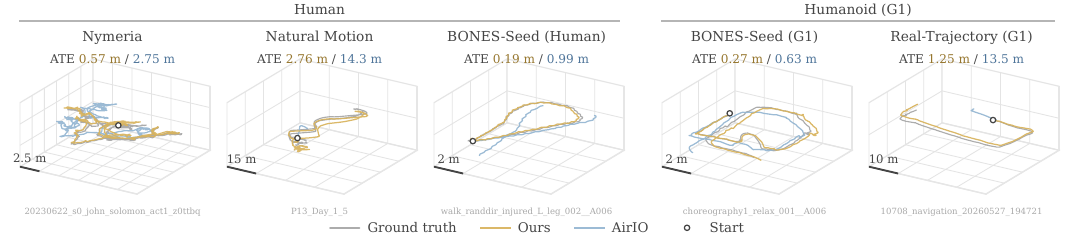}
    \vspace*{-0.8cm}
    \caption{Estimated trajectories sampled from each dataset. Predicted pelvis velocities from our model (gold) and \airio\ (blue) are integrated using ground-truth attitude; gray shows the reference trajectory. Each panel  reports per-recording $\mathrm{ATE}_{\mathrm{GT}}$ in meters (Ours/AirIO).}
    \label{fig:qual_traj}
    \vspace{-16pt}
\end{figure*}

\subsection{Sensor and Motion Randomization}
\label{sec:method_augmentation}
We randomize gravity, sensor noise, bias drift, vibration, and impacts to improve robustness to disturbances absent from kinematic motion. The ranges are chosen heuristically from common sensor specifications and broadened to cover plausible operating conditions (Table~\ref{tab:randomization}).
We also vary body scale, motion speed, and sampling, applying scale and timing changes consistently to inputs and targets while keeping denoising targets free of gravity and sensor disturbances.
Validation omits scale and temporal randomization while retaining noise injection, and real-G1 evaluation uses recorded signals without augmentation.

\section{EXPERIMENTS}
\label{sec:experiments}

We evaluate one jointly trained model on inertial odometry (Sec.~\ref{sec:exp_odometry}) and motion caption retrieval (Sec.~\ref{sec:exp_language}), followed by physical G1 experiments and runtime analysis (Sec.~\ref{sec:exp_realworld}).
We then ablate the training data, input representation, and architecture (Sec.~\ref{sec:exp_ablations}) and analyze compute allocation across model size and compute (Sec.~\ref{sec:ablation_scaling}).

\subsection{Experimental Setup}
\label{sec:exp_setup}

\subsubsection{Datasets}
Training combines \nymeria~\cite{ma2024nymeria}, \naturalmotion~\cite{geissinger2020motion}, and \bones~\cite{bones_studio_2026_seed} human motion with kinematically retargeted \bonesgone\ motion; Table~\ref{tab:datasets} reports total corpus hours before splitting.
Human datasets use an 85\%/15\% training/validation split by trajectory, before window extraction.
For retargeted G1, natural-language descriptions separate simple motions (e.g., walking and standing) from complex motions (e.g., dancing): simple motions follow the same split, while all complex trajectories form the unseen evaluation subset.
Held-out G1 trajectories may have human counterparts in training, testing human-to-humanoid transfer (Sec.~\ref{sec:exp_language}).
All results in this section use held-out validation trajectories or real-G1 recordings, which are reserved entirely for zero-shot evaluation.

\subsubsection{Baselines}
For odometry, we provide the pelvis IMU as input to each single-IMU baseline and compare root-velocity estimates and their integrated trajectories across all methods.
We train \tlio~\cite{liu2020tlio}, \airimu~\cite{qiu2025airimu}, and \airio~\cite{qiu2025airio} per dataset and evaluate the released \tartanimu~\cite{zhao2025tartanimu} human head.
We also augment \airio\ with relative joint positions and velocities to control for the additional information in our synthesized IMUs.
For motion caption retrieval
, we train \imuclip~\cite{moon2023imu2clip} per dataset using the pelvis IMU and evaluate pretrained \imagebind~\cite{girdhar2023imagebind} using the head IMU.

\subsubsection{Implementation Details}
We train for 20 epochs on 4096ms windows with bfloat16 precision and batch size of 1024. Muon~\cite{jordan2024muon} is used to update matrix parameters at learning rate of $2\!\times\!10^{-3}$, while AdamW~\cite{loshchilov2018decoupled} updates the rest at $5\!\times\!10^{-3}$. 
All \textit{Ours} results in this section uses the exponential moving average (EMA)~\cite{morales2024exponential} with weight of $0.95$.

\subsection{Inertial Odometry}
\label{sec:exp_odometry}

\noindent\textbf{Task}\quad We evaluate root-velocity estimation and the resulting integrated trajectories on human and humanoid motion, with representative trajectories in Fig.~\ref{fig:qual_traj}.

\noindent\textbf{Metric}\quad
We evaluate root-motion estimation using absolute velocity error (AVE, m/s) and absolute trajectory error (ATE, m), pooling pelvis samples across recordings.
AVE measures the mean Euclidean error of body-frame velocity predictions:
\begin{equation}
    \mathrm{AVE} = \frac{1}{T}\sum_{t=1}^{T}
    \norm{\hat{\mathbf{v}}_t-\mathbf{v}_t}_2.
\end{equation}
To evaluate accumulated trajectory error, we rotate predicted velocities into the world frame using attitude $\mathbf{R}_t$ and integrate from each recording's ground-truth initial position:
\begin{equation}
\begin{aligned}
    \hat{\mathbf{p}}_t &= \mathbf{p}_0
    + \int_0^t \mathbf{R}_\tau\hat{\mathbf{v}}_\tau\,d\tau, \\
    \mathrm{ATE} &= \sqrt{\frac{1}{T}\sum_{t=1}^{T}
    \norm{\hat{\mathbf{p}}_t-\mathbf{p}_t}_2^2}.
\end{aligned}
\end{equation}
We report $\mathrm{ATE}_{\mathrm{GT}}$ using ground-truth attitude to isolate velocity-estimation error, and $\mathrm{ATE}_{\mathrm{Pred}}$ using gyro-integrated attitude to additionally capture orientation drift.
Gyroscope integration starts from the ground-truth initial attitude; neither setting uses filter updates or trajectory alignment.

\noindent\textbf{Analysis}\quad Table~\ref{tab:odometry} shows that our model achieves the lowest AVE and $\mathrm{ATE}_{\mathrm{GT}}$ on all five datasets, including 0.024~m/s AVE on retargeted G1 and 0.099~m/s on real G1.
It also leads $\mathrm{ATE}_{\mathrm{Pred}}$ on four datasets, with \tlio\ slightly better on \nymeria.
The higher errors under integrated attitude indicate that orientation drift remains a limitation.

\subsection{IMU--Language Retrieval}
\label{sec:exp_language}
\begin{table}[t]
\caption{
Motion caption retrieval:
motion-to-text recall at 1 (R@1, $\uparrow$) / text-to-motion mean average precision (mAP \%, $\uparrow$). 
Unseen G1 categories occur only on human embodiment in training dataset, showing the cross-embodiment knowledge transfer. 
}
\vspace{-6pt}
\label{tab:language}
\centering
\setlength{\tabcolsep}{2pt}
\resizebox{\columnwidth}{!}{%
\begin{tabular}{l rcl rcl rcl rcl}
\toprule
 & \multicolumn{6}{c}{Human} & \multicolumn{6}{c}{Humanoid (G1)} \\
\cmidrule(lr){2-7}\cmidrule(lr){8-13}
Method & \multicolumn{3}{c}{\nymeria} & \multicolumn{3}{c}{\bones} & \multicolumn{3}{c}{Seen cat.} & \multicolumn{3}{c}{Unseen cat.} \\
\midrule
Collapse & 4.4 & / & 0.1 & 19.7 & / & 0.2 & 21.9 & / & 0.8 & 9.8 & / & 0.1 \\
\midrule
\imagebind~\cite{girdhar2023imagebind} & 8.9 & / & 0.1 & 7.1 & / & 0.4 & 11.4 & / & 1.6 & 4.6 & / & 0.1 \\
\imuclip\textsuperscript{$\dagger$}~\cite{moon2023imu2clip} & \best{16.2} & / & 0.4 & 51.8 & / & 5.2 & 36.1 & / & 11.2 & 17.3 & / & 0.5 \\
\specialrule{\lightrulewidth}{\aboverulesep}{0pt}
\rowcolor{oursshade}
\rule[-0.9ex]{0pt}{3.1ex}Ours & 14.9 & / & \best{0.6} & \best{60.7} & / & \best{15.9} & \best{63.4} & / & \best{32.9} & \best{53.8} & / & \best{11.3} \\
\specialrule{\heavyrulewidth}{0pt}{\belowrulesep}
\end{tabular}
}
\par\vspace{2pt}
{\scriptsize\raggedright \textsuperscript{$\dagger$}Trained separately on each dataset.\par}
\vspace*{-16pt}
\end{table}

\begin{figure}[t]
    \centering
    \includegraphics[width=\columnwidth]{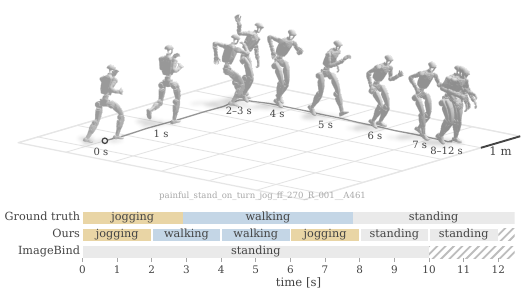}
    \vspace*{-0.6cm}
    \caption{Motion-description retrieval on a retargeted G1 sequence. Sampled poses (top) accompany annotated categories and descriptions retrieved by our model and ImageBind (bottom). Our model uses 2s and ImageBind uses 10s windows; hatching marks intervals without a complete input window.}
    \label{fig:qual_lang}
    \vspace{-10pt}
\end{figure}

\noindent\textbf{Task}\quad We evaluate motion--language alignment in both directions: retrieving descriptions from IMU windows and retrieving motion windows from text queries.
Fig.~\ref{fig:qual_lang} illustrates retrieval over a motion sequence.

\noindent\textbf{Metric}\quad
We rank retrieval candidates by cosine similarity within each dataset.
For IMU-to-text retrieval, we report recall at 1 (R@1) against 19 random distractors averaged over eight draws.
For text-to-motion retrieval, we report mean average precision (mAP) over all annotated motion windows, averaged across distinct text queries.
To assess whether retrieval reflects motion-specific information, we include a \textit{Collapse} baseline that ignores the IMU input and assigns every window the same embedding: the normalized mean of the evaluation text targets.
This baseline measures performance attainable from a constant representation, providing a reference for interpreting retrieval scores when the embedding does not distinguish between motions.

\noindent\textbf{Analysis}\quad Table~\ref{tab:language} shows that our model leads both metrics on \bones\ human motion and both G1 splits.
On categories excluded from G1 training, it reaches 53.8\% R@1 and 11.3\% mAP versus 17.3\% and 0.5\% for \imuclip, supporting semantic transfer from human motion.
On \nymeria, \imuclip\ achieves higher R@1 (16.2\% versus 14.9\%), while our model achieves higher mAP.

\subsection{Real-world Experiment and Runtime Analysis}
\label{sec:exp_realworld}
\begin{figure}[t]
    \centering
    \includegraphics[width=\columnwidth]{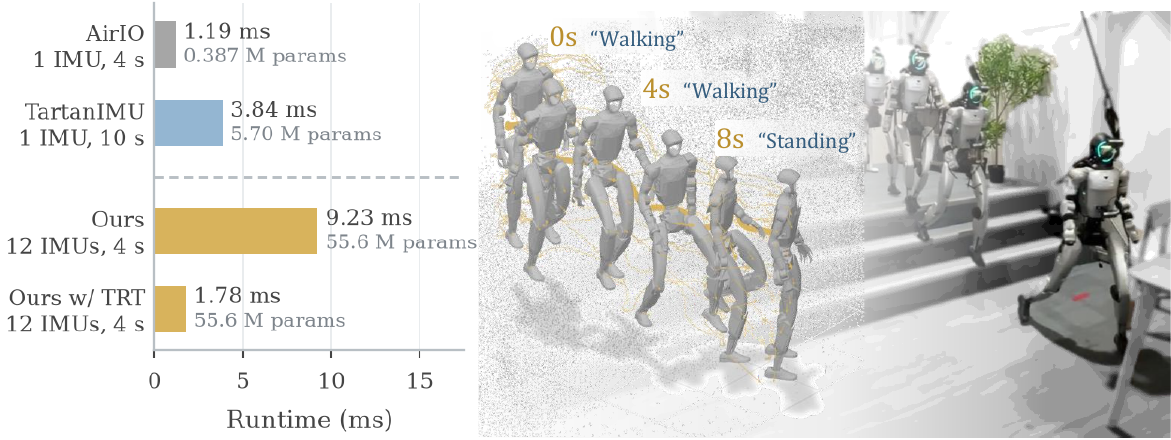}
    \vspace{-14pt}
    \caption{Inference runtime and deployment on a physical G1. Left: latency on an NVIDIA L40S with 200-Hz inputs; labels show IMU count, window duration, and model size. TRT denotes TensorRT. Right: stair descent shown by robot poses over a 3D point cloud, with retrieved descriptions changing from ``Walking'' to ``Standing'' as the robot stops.}
    \label{fig:realworld}
    \vspace{-16pt}
\end{figure}

We evaluate motion tracking and language retrieval on a physical G1 without real-robot training or fine-tuning.
Fig.~\ref{fig:realworld} shows the tracked stair descent, with retrieved descriptions changing from ``Walking'' to ``Standing'' as the robot stops.

Despite its greater capacity and joint support for state estimation and motion understanding, our encoder remains suitable for real-time robot use. With TensorRT on an NVIDIA L40S, it processes a 4-s window of 12 IMUs sampled at 200~Hz in 1.78~ms (Fig.~\ref{fig:realworld}). This latency is well below the 5-ms input sampling interval, allowing inference to keep pace with incoming measurements. 

\subsection{Ablation Studies}
\label{sec:exp_ablations}
\begin{table}[t]
\caption{Model and training recipe ablations (ATE in m, lower is better), grouped by the four questions in Sec.~\ref{sec:exp_ablations}.}
\vspace{-6pt}
\label{tab:ablation_multiimu_scaling}
\centering
\setlength{\tabcolsep}{3pt}
\resizebox{\columnwidth}{!}{%
\begin{tabular}{lcccccc}
\toprule
 & \multicolumn{2}{c}{\nymeria} & \multicolumn{2}{c}{\bonesgone} & \multicolumn{2}{c}{\superodom} \\
\cmidrule(lr){2-3}\cmidrule(lr){4-5}\cmidrule(lr){6-7}
Variant & ATE\textsubscript{GT} & ATE\textsubscript{Pred} & ATE\textsubscript{GT} & ATE\textsubscript{Pred} & ATE\textsubscript{GT} & ATE\textsubscript{Pred} \\
\midrule
\multicolumn{7}{l}{\textit{1. Training embodiment}} \\
\hspace{5mm}Humanoid-only & 8.041 & 17.50 & 0.162 & 0.180 & 5.901 & 13.15 \\
\midrule
\multicolumn{7}{l}{\textit{2. Input representation}} \\
\hspace{5mm}Single pelvis IMU & 3.592 & 16.66 & 0.330 & 0.335 & 8.114 & 12.81 \\
\hspace{5mm}Pelvis IMU + joints & \best{2.283} & 16.12 & 0.136 & 0.164 & 3.513 & 18.36 \\
\midrule
\multicolumn{7}{l}{\textit{3. Model design}} \\
\hspace{5mm}w/o Alternating Attn. & 2.466 & \best{14.31} & 0.110 & 0.128 & 3.530 & 11.71 \\
\hspace{5mm}w/o Continuous Enc.    & 31.42 & 31.51 & 1.189 & 1.189 & 17.37 & 17.67 \\
\midrule
\multicolumn{7}{l}{\textit{4. Data diversity}} \\
\hspace{5mm}1/64 of data & 4.882 & 18.40 & 0.154 & 0.166 & 4.369 & 12.44 \\
\hspace{5mm}1/16 of data & 3.500 & 16.93 & 0.111 & 0.131 & 3.792 & 11.95 \\
\hspace{5mm}1/4 of data  & 2.847 & 14.64 & \best{0.099} & \best{0.121} & 3.716 & 11.98 \\
\specialrule{\lightrulewidth}{\aboverulesep}{0pt}
\rowcolor{oursshade}
\rule[-0.9ex]{0pt}{3.1ex}%
Ours & 2.468 & 18.20 & 0.101 & 0.127 & \best{3.153} & \best{11.50} \\
\specialrule{\heavyrulewidth}{0pt}{\belowrulesep}
\end{tabular}}
\vspace{-10pt}
\end{table}

We examine how human motion data, the shared multi-IMU representation, encoder design, and training-data diversity contribute to motion estimation across embodiments. Table~\ref{tab:ablation_multiimu_scaling} shows the results on human and humanoid datasets.

\subsubsection{Human-to-Humanoid Transfer}
\label{sec:ablation_transfer}
\begin{figure}[t]
    \centering
    \includegraphics[width=\linewidth]{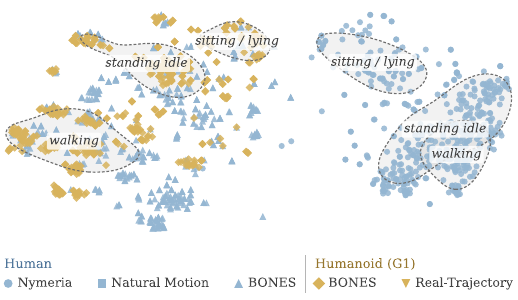}
    \vspace{-14pt}
    \caption{Human and humanoid motion in the learned feature space. A t-SNE projection of \texttt{[CLS]} embeddings uses blue for humans, gold for G1, and marker shapes for datasets. Dashed regions enclose 60\% of windows in each motion category, showing overlap between human and humanoid data.}
    \label{fig:tsne}
    \vspace{-16pt}
\end{figure}

Diverse human motion may provide useful motion priors beyond those learned from humanoid data alone. We compare our jointly trained model with a humanoid specialist trained exclusively on retargeted G1 motion. Joint training improves both trajectory metrics on retargeted and real G1, reducing $\mathrm{ATE}_{\mathrm{GT}}$ from 0.162 to 0.101~m and from 5.901 to 3.153~m, respectively (Table~\ref{tab:ablation_multiimu_scaling}). These results suggest that broader human motion experience can benefit humanoid odometry even relative to embodiment-specific training. Fig.~\ref{fig:tsne} further illustrates the shared representation, with corresponding human and G1 motion categories overlapping in the t-SNE plot.

\subsubsection{Multi-IMU Representation}
\label{sec:ablation_multiimu}
Our shared inertial interface incorporates embodiment geometry through kinematic IMU synthesis, aligning human and humanoid motion at corresponding body sites. We compare against pelvis-only input to assess the value of whole-body motion cues, and against pelvis IMU plus joint positions and velocities to test their representation in a common inertial space. 
Table~\ref{tab:ablation_multiimu_scaling} shows that multi-IMU input reduces real-G1 $\mathrm{ATE}_{\mathrm{GT}}$ from 8.114 to 3.153~m relative to pelvis-only input and improves both trajectory metrics over direct joints on retargeted and real G1. Although direct joints perform better on \nymeria, the humanoid results support incorporating embodiment geometry into a common inertial representation.

\subsubsection{Model Design}
\label{sec:ablation_modeldesign}
We examine whether explicitly modeling each site's temporal structure and physical sampling time benefits the shared encoder (Table~\ref{tab:ablation_multiimu_scaling}). Alternating within-site and full attention modestly improves humanoid odometry over full attention alone, suggesting that dedicated per-site modeling complements cross-body interactions, although gains on \nymeria\ are mixed. Continuous-time tokenization yields larger gains across all three datasets over fixed-sample patches, whose duration varies with sampling rate, supporting a consistent temporal representation across embodiments.

\subsubsection{Data Diversity}
\label{sec:ablation_data_diversity}
We test whether broader motion coverage improves generalization by varying unique training sequences at fixed model size and approximately matched optimization budgets (Table~\ref{tab:ablation_multiimu_scaling}). More diverse data consistently reduces $\mathrm{ATE}_{\mathrm{GT}}$ on \nymeria\ and real G1, supporting broader motion exposure over repeated training on a smaller corpus. Retargeted G1 performs best with quarter-data training, suggesting that additional diversity can benefit hardware transfer without improving retargeted-motion accuracy.

\subsection{Compute Allocation}
\label{sec:ablation_scaling}
Efficient scaling requires balancing model capacity with training exposure under a limited compute budget. To characterize this tradeoff, we fit a Chinchilla-style scaling law~\cite{hoffmann2022chinchilla} across models of varying width and depth:
\begin{equation}
    L(N,D)=E+\frac{A}{N^{\alpha}}+\frac{B}{D^{\beta}},
    \label{eq:scaling_law}
\end{equation}
where $L$ is shared-site validation velocity error, $N$ is the parameter count, and $D$ is the number of processed tokens, including repeated windows. The fit estimates an error floor $E$, coefficients $A,B$, and scaling exponents $\alpha,\beta$.

\begin{figure}[t]
    \centering
    \includegraphics[width=\columnwidth]{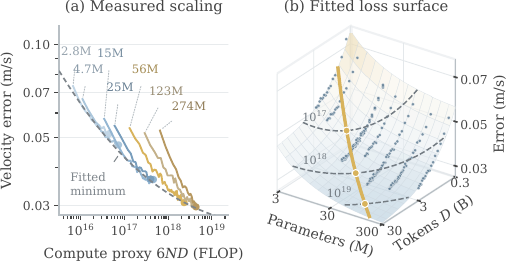}
    \vspace{-14pt}
    \caption{Model capacity and training exposure under a fixed compute budget. (a) Measured velocity error and fitted compute-optimal trend (dashed). (b) Fitted error surface with observations (blue), constant-$6ND$ contours (gray), and predicted optima (gold). $N$ counts parameters and $D$ counts processed tokens, including repeats; the surface includes extrapolation. 
    }
    \label{fig:scaling}
    \vspace{-16pt}
\end{figure}

Under the compute proxy $C=6ND$, the fit yields $N_{\mathrm{opt}}\propto C^{0.52}$ and $D_{\mathrm{opt}}\propto C^{0.48}$ (Fig.~\ref{fig:scaling}). These nearly equal exponents suggest that increasing model capacity and training exposure together uses additional compute effectively. The fitted contours illustrate this tradeoff and guide allocation across compute budgets. These estimates include extrapolation and optimize validation error, so they need not yield the best hardware transfer.

\section{CONCLUSIONS}
\label{sec:conclusion}

We presented a unified inertial model that zero-shot transfers human motion knowledge to a physical humanoid for motion estimation and language understanding, without physics simulation, real-robot training trajectories, or fine-tuning. A shared inertial interface incorporates embodiment geometry through kinematic IMU synthesis, while continuous-time encoding accommodates different sampling rates. Together, these designs enable a single checkpoint to achieve the lowest velocity error among the evaluated methods across five datasets. The learned language-aligned representation also supports motion interpretation through text retrieval without task-specific classifiers, including humanoid motion categories absent from G1 training.

Hardware evaluation currently covers a single humanoid platform, and language supervision is partly synthetic with a limited vocabulary. Despite accurate velocity estimates, orientation drift remains a major source of trajectory error. Future work will extend evaluation across humanoid platforms and integrate learned velocities with filtering for closed-loop state estimation and control.

\bibliographystyle{IEEEtran}
\bibliography{references}

@article{liu2020tlio,
  author  = {Liu, Wenxin and Caruso, David and Ilg, Eddy and Dong, Jing and Mourikis, Anastasios I. and Daniilidis, Kostas and Kumar, Vijay and Engel, Jakob},
  title   = {{TLIO}: Tight Learned Inertial Odometry},
  journal = {IEEE Robotics and Automation Letters},
  volume  = {5},
  number  = {4},
  pages   = {5653--5660},
  year    = {2020}
}

@article{qiu2025airimu,
  author  = {Qiu, Yuheng and Wang, Chen and Xu, Can and Chen, Yutian and Zhou, Xunfei and Xia, Youjie and Scherer, Sebastian},
  title   = {{AirIMU}: Learning Uncertainty Propagation for Inertial Odometry},
  journal = {arXiv preprint arXiv:2310.04874},
  year    = {2023}
}

@article{qiu2025airio,
  title = {{AirIO}: Learning Inertial Odometry With Enhanced {IMU} Feature Observability},
  author = {Qiu, Yuheng and Xu, Can and Chen, Yutian and Zhao, Shibo and Geng, Junyi and Scherer, Sebastian},
  journal = {IEEE Robotics and Automation Letters},
  year = {2025},
  volume = {10},
  number = {9},
  pages = {9368--9375},
  doi = {10.1109/LRA.2025.3581130},
  url = {https://doi.org/10.1109/LRA.2025.3581130}
}

@inproceedings{zhao2025tartanimu,
  author    = {Zhao, Shibo and Zhou, Sifan and Blanchard, Raphael and Qiu, Yuheng and Wang, Wenshan and Scherer, Sebastian},
  title     = {Tartan {IMU}: A Light Foundation Model for Inertial Positioning in Robotics},
  booktitle = {Proceedings of the IEEE/CVF Conference on Computer Vision and Pattern Recognition (CVPR)},
  pages     = {22520--22529},
  year      = {2025}
}

@inproceedings{herath2020ronin,
  author    = {Herath, Sachini and Yan, Hang and Furukawa, Yasutaka},
  title     = {{RoNIN}: Robust Neural Inertial Navigation in the Wild: Benchmark, Evaluations, \& New Methods},
  booktitle = {IEEE International Conference on Robotics and Automation (ICRA)},
  pages     = {3146--3152},
  year      = {2020}
}

@inproceedings{chen2018ionet,
  author    = {Chen, Changhao and Lu, Xiaoxuan and Markham, Andrew and Trigoni, Niki},
  title     = {{IONet}: Learning to Cure the Curse of Drift in Inertial Odometry},
  booktitle = {Proceedings of the AAAI Conference on Artificial Intelligence},
  volume    = {32},
  number    = {1},
  pages     = {6468--6476},
  year      = {2018}
}

@inproceedings{girdhar2023imagebind,
  author    = {Girdhar, Rohit and El-Nouby, Alaaeldin and Liu, Zhuang and Singh, Mannat and Alwala, Kalyan Vasudev and Joulin, Armand and Misra, Ishan},
  title     = {{ImageBind}: One Embedding Space To Bind Them All},
  booktitle = {Proceedings of the IEEE/CVF Conference on Computer Vision and Pattern Recognition (CVPR)},
  pages     = {15180--15190},
  year      = {2023}
}

@inproceedings{moon2023imu2clip,
  author    = {Moon, Seungwhan and Madotto, Andrea and Lin, Zhaojiang and Saraf, Aparajita and Bearman, Amy and Damavandi, Babak},
  title     = {{IMU2CLIP}: Language-grounded Motion Sensor Translation with Multimodal Contrastive Learning},
  booktitle = {Findings of the Association for Computational Linguistics: EMNLP 2023},
  pages     = {13246--13253},
  year      = {2023}
}

@inproceedings{das2025primus,
  title = {{PRIMUS}: Pretraining {IMU} Encoders with Multimodal Self-Supervision},
  author = {Das, Arnav M. and Tang, Chi Ian and Kawsar, Fahim and Malekzadeh, Mohammad},
  booktitle = {ICASSP 2025 -- 2025 IEEE International Conference on Acoustics, Speech and Signal Processing (ICASSP)},
  year = {2025},
  pages = {1--5},
  publisher = {IEEE},
  doi = {10.1109/ICASSP49660.2025.10888874},
  url = {https://doi.org/10.1109/ICASSP49660.2025.10888874}
}

@inproceedings{xu2021limubert,
  author    = {Xu, Huatao and Zhou, Pengfei and Tan, Rui and Li, Mo and Shen, Guobin},
  title     = {{LIMU-BERT}: Unleashing the Potential of Unlabeled Data for {IMU} Sensing Applications},
  booktitle = {Proceedings of the 19th ACM Conference on Embedded Networked Sensor Systems (SenSys)},
  year      = {2021}
}

@inproceedings{zhang2025sensorlm,
  author    = {Zhang, Yuwei and Ayush, Kumar and Qiao, Siyuan and Heydari, A. Ali and Narayanswamy, Girish and Xu, Maxwell A. and Metwally, Ahmed A. and Xu, Shawn and Garrison, Jake and Xu, Xuhai and Althoff, Tim and Liu, Yun and Kohli, Pushmeet and Zhan, Jiening and Malhotra, Mark and Patel, Shwetak and Mascolo, Cecilia and Liu, Xin and McDuff, Daniel and Yang, Yuzhe},
  title     = {{SensorLM}: Learning the Language of Wearable Sensors},
  booktitle = {Advances in Neural Information Processing Systems},
  volume    = {38},
  year      = {2025}
}

@article{matsuishi2025aura,
  author  = {Matsuishi, Koki and Ukita, Kosuke and Okita, Tsuyoshi},
  title   = {Multimodal Foundation Model for Cross-Modal Retrieval and Activity Recognition Tasks},
  journal = {International Journal of Activity and Behavior Computing},
  volume  = {2025},
  number  = {3},
  pages   = {1--25},
  doi     = {10.60401/ijabc.124},
  year    = {2025}
}

@inproceedings{choi2020retargeting,
  author    = {Choi, Sungjoon and Pan, Matthew and Kim, Joohyung},
  title     = {Nonparametric Motion Retargeting for Humanoid Robots on Shared Latent Space},
  booktitle = {Robotics: Science and Systems},
  year      = {2020}
}

@inproceedings{he2024h2o,
  title = {Learning Human-to-Humanoid Real-Time Whole-Body Teleoperation},
  author = {He, Tairan and Luo, Zhengyi and Xiao, Wenli and Zhang, Chong and Kitani, Kris and Liu, Changliu and Shi, Guanya},
  booktitle = {2024 IEEE/RSJ International Conference on Intelligent Robots and Systems (IROS)},
  year = {2024},
  publisher = {IEEE},
  doi = {10.1109/IROS58592.2024.10801984},
  url = {https://doi.org/10.1109/IROS58592.2024.10801984}
}

@inproceedings{he2024omnih2o,
  title = {{OmniH2O}: Universal and Dexterous Human-to-Humanoid Whole-Body Teleoperation and Learning},
  author = {He, Tairan and Luo, Zhengyi and He, Xialin and Xiao, Wenli and Zhang, Chong and Zhang, Weinan and Kitani, Kris M. and Liu, Changliu and Shi, Guanya},
  booktitle = {Proceedings of The 8th Conference on Robot Learning},
  year = {2025},
  volume = {270},
  series = {Proceedings of Machine Learning Research},
  pages = {1516--1540},
  publisher = {PMLR},
  url = {https://proceedings.mlr.press/v270/he25b.html}
}

@inproceedings{fu2024humanplus,
  title = {{HumanPlus}: Humanoid Shadowing and Imitation from Humans},
  author = {Fu, Zipeng and Zhao, Qingqing and Wu, Qi and Wetzstein, Gordon and Finn, Chelsea},
  booktitle = {Proceedings of The 8th Conference on Robot Learning},
  year = {2025},
  volume = {270},
  series = {Proceedings of Machine Learning Research},
  pages = {2828--2844},
  publisher = {PMLR},
  url = {https://proceedings.mlr.press/v270/fu25a.html}
}

@article{huang2026human2humanoid,
  author  = {Huang, Tianchen and Yuan, Feiyang and Gu, Junchi and Fang, Shurui and Zhang, Xiaohu and Wang, Yu and Gao, Wei and Zhang, Shiwu},
  title   = {{Human2Humanoid}: Physics-Aware Cross-Morphology Motion Retargeting for Humanoid Robots},
  journal = {arXiv preprint arXiv:2606.03476},
  year    = {2026}
}

@inproceedings{ma2024nymeria,
  author    = {Ma, Lingni and Ye, Yuting and Hong, Fangzhou and Guzov, Vladimir and Jiang, Yifeng and Postyeni, Rowan and Pesqueira, Luis and Gamino, Alexander and Baiyya, Vijay and Kim, Hyo Jin and Bailey, Kevin and Fosas, David Soriano and Liu, C. Karen and Liu, Ziwei and Engel, Jakob and De Nardi, Renzo and Newcombe, Richard},
  title     = {Nymeria: A Massive Collection of Multimodal Egocentric Daily Motion in the Wild},
  booktitle = {European Conference on Computer Vision (ECCV)},
  year      = {2024}
}

@article{geissinger2020motion,
  author  = {Geissinger, Jack H. and Asbeck, Alan T.},
  title   = {Motion Inference Using Sparse Inertial Sensors, Self-Supervised Learning, and a New Dataset of Unscripted Human Motion},
  journal = {Sensors},
  volume  = {20},
  number  = {21},
  pages   = {6330},
  year    = {2020}
}

@misc{bones_studio_2026_seed,
  title = {{BONES-SEED}: Skeletal Everyday Embodied Dataset},
  author = {{Bones Studio}},
  year = {2026},
  url = {https://huggingface.co/datasets/bones-studio/seed},
  note = {Accessed: 2026-08-16}
}

@inproceedings{hoffmann2022chinchilla,
  author    = {Hoffmann, Jordan and Borgeaud, Sebastian and Mensch, Arthur and Buchatskaya, Elena and Cai, Trevor and Rutherford, Eliza and de Las Casas, Diego and Hendricks, Lisa Anne and Welbl, Johannes and Clark, Aidan and Hennigan, Thomas and Noland, Eric and Millican, Katherine and van den Driessche, George and Damoc, Bogdan and Guy, Aurelia and Osindero, Simon and Simonyan, Kar\'{e}n and Elsen, Erich and Vinyals, Oriol and Rae, Jack and Sifre, Laurent},
  title     = {An Empirical Analysis of Compute-Optimal Large Language Model Training},
  booktitle = {Advances in Neural Information Processing Systems},
  volume    = {35},
  pages     = {30016--30030},
  year      = {2022},
  doi       = {10.52202/068431-2176}
}

@inproceedings{radford2021clip,
  author    = {Radford, Alec and Kim, Jong Wook and Hallacy, Chris and Ramesh, Aditya and Goh, Gabriel and Agarwal, Sandhini and Sastry, Girish and Askell, Amanda and Mishkin, Pamela and Clark, Jack and Krueger, Gretchen and Sutskever, Ilya},
  title     = {Learning Transferable Visual Models From Natural Language Supervision},
  booktitle = {Proceedings of the 38th International Conference on Machine Learning (ICML)},
  series    = {Proceedings of Machine Learning Research},
  volume    = {139},
  pages     = {8748--8763},
  year      = {2021}
}

@inproceedings{rotella2014state,
  author    = {Rotella, Nicholas and Bloesch, Michael and Righetti, Ludovic and Schaal, Stefan},
  title     = {State Estimation for a Humanoid Robot},
  booktitle = {IEEE/RSJ International Conference on Intelligent Robots and Systems (IROS)},
  pages     = {952--958},
  year      = {2014},
  doi       = {10.1109/IROS.2014.6942674}
}

@article{kuindersma2016atlas,
  author  = {Kuindersma, Scott and Deits, Robin and Fallon, Maurice and Valenzuela, Andr{\'e}s and Dai, Hongkai and Permenter, Frank and Koolen, Twan and Marion, Pat and Tedrake, Russ},
  title   = {Optimization-based locomotion planning, estimation, and control design for the {Atlas} humanoid robot},
  journal = {Autonomous Robots},
  volume  = {40},
  number  = {3},
  pages   = {429--455},
  year    = {2016},
  doi     = {10.1007/s10514-015-9479-3}
}

@inproceedings{hayes2017transparency,
  author    = {Hayes, Bradley and Shah, Julie A.},
  title     = {Improving Robot Controller Transparency Through Autonomous Policy Explanation},
  booktitle = {ACM/IEEE International Conference on Human-Robot Interaction (HRI)},
  pages     = {303--312},
  year      = {2017},
  doi       = {10.1145/2909824.3020233}
}

@article{fallon2018localization,
  author  = {Fallon, Maurice},
  title   = {Accurate and robust localization for walking robots fusing kinematics, inertial, vision and {LIDAR}},
  journal = {Interface Focus},
  volume  = {8},
  number  = {4},
  pages   = {20180015},
  year    = {2018},
  doi     = {10.1098/rsfs.2018.0015}
}

@inproceedings{cao2022rio,
  author = {Cao, Xiya and Zhou, Caifa and Zeng, Dandan and Wang, Yongliang},
  title = {{RIO}: Rotation-Equivariance Supervised Learning of Robust Inertial Odometry},
  booktitle = {Proceedings of the IEEE/CVF Conference on Computer Vision and Pattern Recognition (CVPR)},
  year = {2022}
}

@inproceedings{jayanth2025eqnio,
  author = {Jayanth, Royina Karegoudra and Xu, Yinshuang and Wang, Ziyun and Chatzipantazis, Evangelos and Daniilidis, Kostas and Gehrig, Daniel},
  title = {{EqNIO}: Subequivariant Neural Inertial Odometry},
  booktitle = {International Conference on Learning Representations (ICLR)},
  year = {2025}
}

@manual{xsens_mvn_manual,
  title = {{MVN} User Manual},
  organization = {{Xsens}},
  url = {https://www.xsens.com/hubfs/Downloads/Manuals/MVN_User_Manual.pdf},
  note = {Sections on coordinate systems and segment axes. Accessed: 2026-09-10}
}

@inproceedings{romero2022ckconv,
  author = {Romero, David W. and Kuzina, Anna and Bekkers, Erik J. and Tomczak, Jakub M. and Hoogendoorn, Mark},
  title = {{CKConv}: Continuous Kernel Convolution for Sequential Data},
  booktitle = {International Conference on Learning Representations (ICLR)},
  year = {2022},
  url = {https://openreview.net/forum?id=8FhxBtXSl0}
}

@article{forster2017preintegration,
  author = {Forster, Christian and Carlone, Luca and Dellaert, Frank and Scaramuzza, Davide},
  title = {On-Manifold Preintegration for Real-Time Visual--Inertial Odometry},
  journal = {IEEE Transactions on Robotics},
  volume = {33},
  number = {1},
  pages = {1--21},
  year = {2017},
  doi = {10.1109/TRO.2016.2597321}
}

@inproceedings{dao2023flashattention2,
  title={Flash{A}ttention-2: Faster Attention with Better Parallelism and Work Partitioning},
  author={Dao, Tri},
  booktitle={International Conference on Learning Representations (ICLR)},
  year={2024}
}

@misc{jordan2024muon,
  author = {Keller Jordan and Yuchen Jin and Vlado Boza and Jiacheng You and Franz Cesista and Laker Newhouse and Jeremy Bernstein},
  title = {Muon: An optimizer for hidden layers in neural networks},
  url = {https://kellerjordan.github.io/posts/muon/},
  year = {2024}
}

@inproceedings{loshchilov2018decoupled,
  title = {Decoupled Weight Decay Regularization},
  author = {Loshchilov, Ilya and Hutter, Frank},
  booktitle = {International Conference on Learning Representations},
  year = {2019},
  url = {https://openreview.net/forum?id=Bkg6RiCqY7}
}

@article{morales2024exponential,
  title = {Exponential Moving Average of Weights in Deep Learning: Dynamics and Benefits},
  author = {Morales-Brotons, Daniel and Vogels, Thijs and Hendrikx, Hadrien},
  journal = {Transactions on Machine Learning Research},
  year = {2024},
  url = {https://openreview.net/forum?id=2M9CUnYnBA}
}

@article{su2024roformer,
  title = {{RoFormer}: Enhanced transformer with {Rotary Position Embedding}},
  author = {Su, Jianlin and Ahmed, Murtadha and Lu, Yu and Pan, Shengfeng and Bo, Wen and Liu, Yunfeng},
  journal = {Neurocomputing},
  volume = {568},
  pages = {127063},
  year = {2024},
  doi = {10.1016/j.neucom.2023.127063},
  url = {https://www.sciencedirect.com/science/article/pii/S0925231223011864}
}

\end{document}